\documentclass[10pt,twocolumn,letterpaper]{article}

\usepackage{cvpr}
\usepackage{times}
\usepackage{epsfig}
\usepackage{graphicx}
\usepackage{amsmath}
\usepackage{amssymb}
\usepackage{microtype}
\usepackage{booktabs}
\usepackage{tabularx}
\usepackage{subfig}
\usepackage{xcolor}

\usepackage[pagebackref=true,breaklinks=true,letterpaper=true,colorlinks,bookmarks=false]{hyperref}

\cvprfinalcopy 

\ifcvprfinal\fi

\def\cvprPaperID{1959} 

\ifcvprfinal\fi
\begin{document}

\title{Video Generation from Single Semantic Label Map}

\author{Junting Pan$^{1,3}$, Chengyu Wang$^1$, Xu Jia$^2$, Jing Shao$^1$, Lu Sheng$^3$, Junjie Yan$^1$, and Xiaogang Wang$^{1,3}$\\
\\
Sensetime Research$^1$, Noah's Ark Lab, Huawei$^2$, \\
CUHK-SenseTime Joint Lab, The Chinese University of Hong Kong$^3$\\
\and
}

\maketitle

\begin{abstract}

This paper proposes the novel task of video generation conditioned on a SINGLE semantic label map, which provides a good balance between flexibility and quality in the generation process. Different from typical end-to-end approaches, which model both scene content and dynamics in a single step, we propose to decompose this difficult task into two sub-problems. As current image generation methods do better than video generation in terms of detail, we synthesize high quality content by only generating the first frame. Then we animate the scene based on its semantic meaning to obtain temporally coherent video, giving us excellent results overall.
%
We employ a cVAE for predicting optical flow as a beneficial intermediate step to generate a video sequence conditioned on the initial single frame. A semantic label map is integrated into the flow prediction module to achieve major improvements in the image-to-video generation process. 
Extensive experiments on the Cityscapes dataset show that our method outperforms all competing methods. 
The source code will be released on \href{https://github.com/junting/seg2vid}{https://github.com/junting/seg2vid}.

\end{abstract}›


\section{Introduction}
\label{sec:Introduction}



A typical visual scene is composed of foreground objects and the background. In a dynamic scene, motion of the background is determined by camera movement which is independent of the motion of foreground objects.
Scene understanding, which include both understanding how foreground objects and background look and how they change, is essential to advancing the development of computer vision. Scene understanding, besides using recognition models, can be accomplished by generative methods\cite{vondrick2016generating}. In this work we focus on using generative models to understand our visual world.


There has been much progress in image generation to address static scene modeling. Researchers have proposed methods to generate images from only noise \cite{goodfellow2014generative} or from pre-defined conditions such as attribute, text and pose \cite{zhang2017stackgan, ma2017pose}. In recent works, people also pay attention to image generation conditioned on semantic information with either paired \cite{isola2017image} or unpaired data \cite{zhu2017unpaired}. The conditional image generation methods provide a way to manipulate existing images and have potential value as a data augmentation strategy to assist other computer vision tasks.
\begin{figure}[t]
\includegraphics[width=\linewidth]{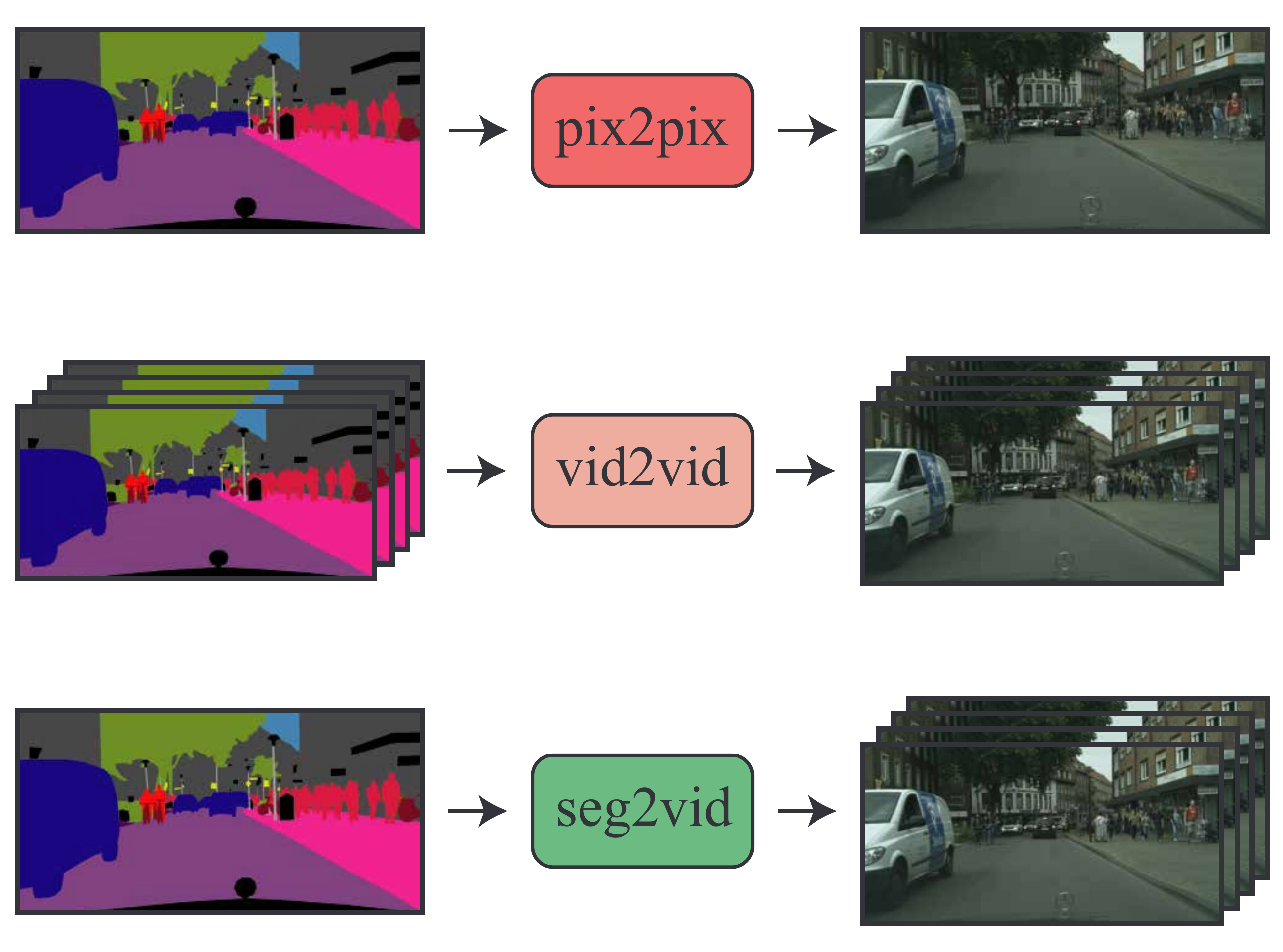}
\caption{Comparison with existing generation tasks. From top: Image-to-image translation, video-to-video, and our image-to-video synthesis. Our method only takes {\it one} semantic label map as input and synthesizes a sequence of photo-realistic video frames. }
\label{fig:figure1}
\end{figure}
While image generation tasks only model static scenes, for video prediction, it is essential to also investigate the temporal dynamics. Models are trained to predict raw pixels of the future frame by learning from historical motion patterns. There is another line of work on video synthesis without any history frames.


Similar to research on image generation, some work investigated unconditional video generation. That is, directly generating video clips from noise by using generative adversarial networks to learn a mapping between spatial-temporal latent space and video clips \cite{tulyakov2017mocogan, saito2017temporal}. Another group of researchers worked on video-to-video translation \cite{wang2018video}, where a sequence of frames are generated according to a sequence of aligned semantic representations.

\begin{figure}[t]
\includegraphics[width=\linewidth]{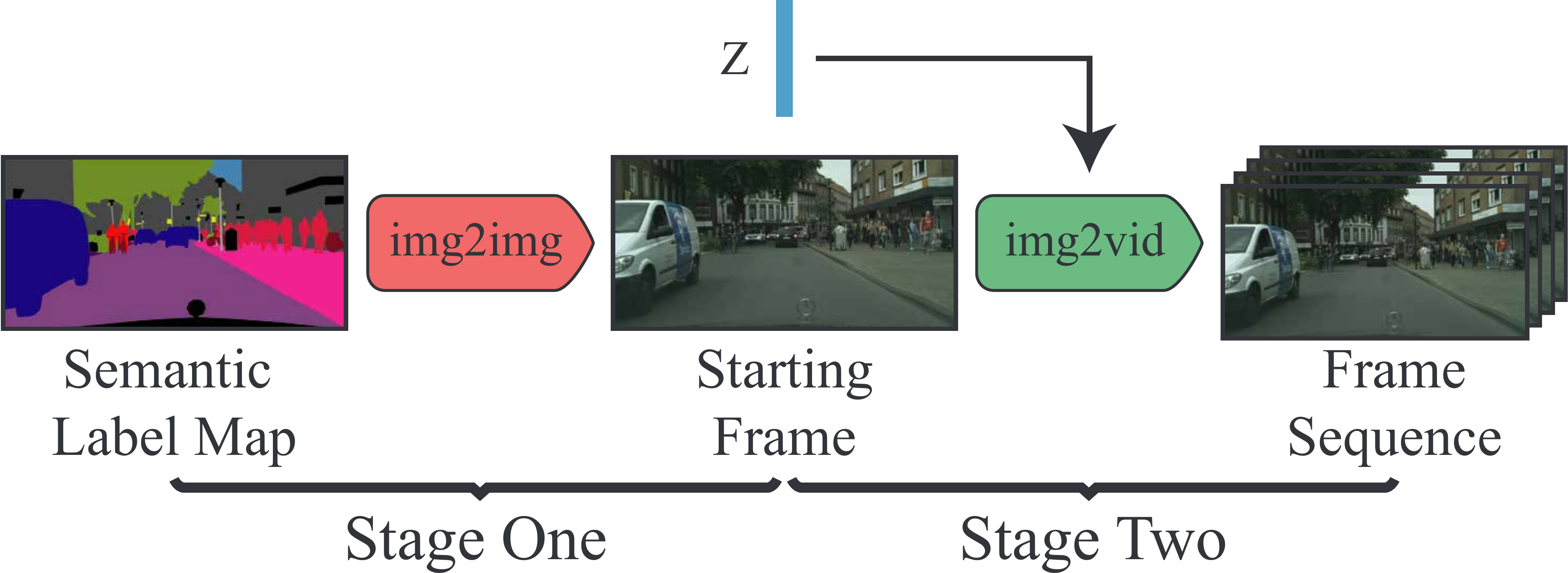}
\caption{Overview of our two step generation network. In the first stage, we generate the starting frame by mapping from a semantic label map. In the second stage, we use our flow prediction network to transform the initial frame to a video sequence.}

\label{fig:overview}
\end{figure}

In this work, we study video generation with a setting similar to the video-to-video work \cite{wang2018video} except that it is only conditioned on a single frame`s semantic label map. 
Compared to previous works on video generation, our setting not only provides control over the generation process but also allows high variability in the results.
Conditioning the generation on semantic label map helps avoid producing undesirable results (\eg~ a car driving on the pavement) which often occurs in unconditional generation. Furthermore, we can generate cars moving at different speeds or in different directions, which is not possible in the video-to-video setting.  
\begin{figure*}[!ht]
\includegraphics[width=\textwidth]{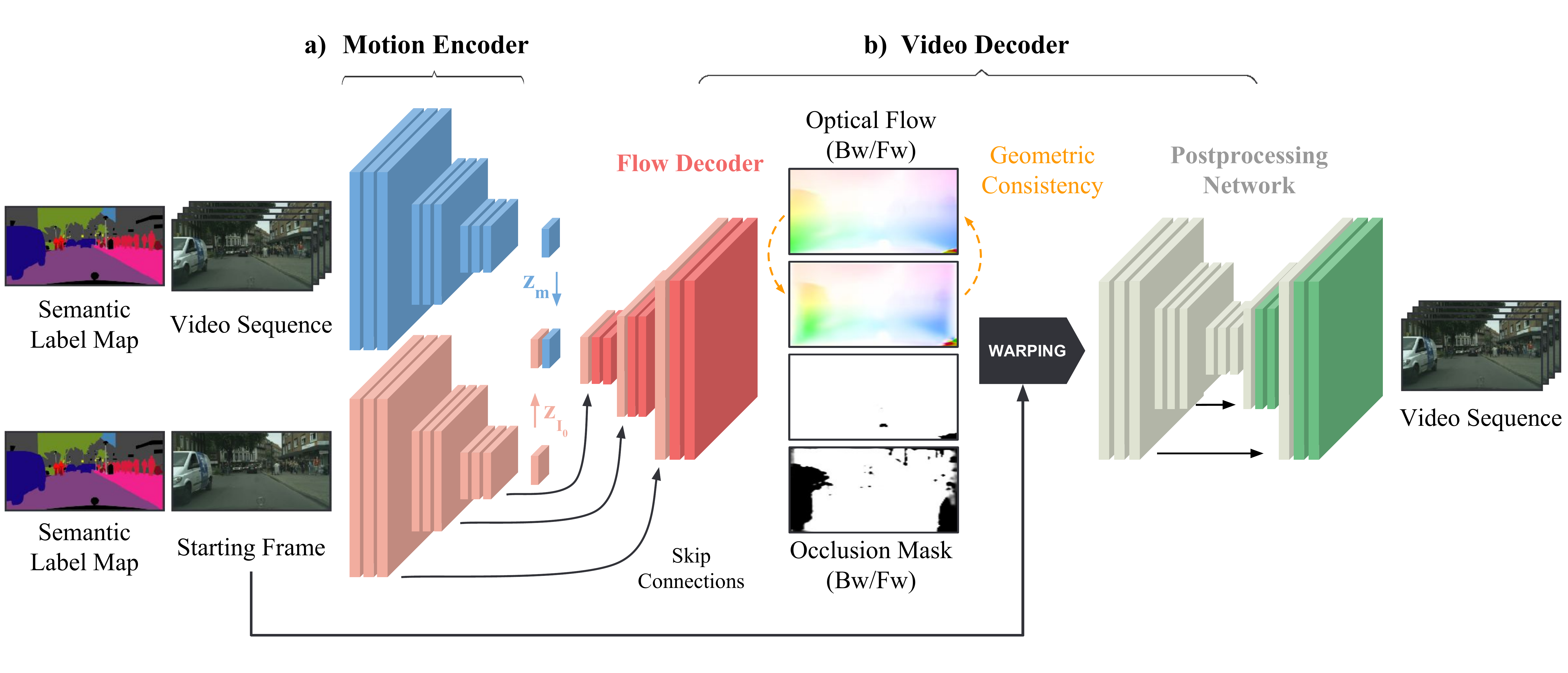}
\caption{Overall architecture of the proposed image-to-video generation network. It consists two components: a) Motion Encoder and b) Video Decoder. For any pair of bidirectional flow predictions, consistency check is computed only in non occluded areas.}
\label{fig:architecture}
\end{figure*}
One intuitive idea to address this new task would be to train an end-to-end conditional generative model. However, it is not easy to apply such a model to datasets composed of diverse objects and background, ~\ie
different objects in different scenes have different motions. In reality, training a single end-to-end model to simultaneously model both appearance and motion of these objects and scenes is very hard. Therefore, as illustrated in Fig. \ref{fig:overview}, we take a divide-and-conquer strategy, designed to model appearance and motion in a progressive manner. 

In the first stage, we aim to transform a semantic label map to a frame such that the appearance of scene is synthesized, which falls into the category of image-to-image translation. During translation process, the model only focuses on producing an image of good quality with reasonable content.

In the next stage, future motion of the scene is predicted based on the generated frame. Specifically, a conditional VAE is employed to model uncertainty of future motion. 
Different from existing video prediction tasks where motion information can be estimated from historical frames, in our setting, we only have one semantic label map and one generated frame available. We argue that it is important for the model to leverage the semantic meaning of the first frame when predicting motion. For example, buildings and pedestrians have very distinctive motion. We take both the semantic label map and the generated frame as input and feed them into a motion prediction model. Empirical results demonstrate that with semantic representation as input, the model can learn better motion for dynamic objects than without that, specially for complex scenes with multiple classes of objects.
We model motion with optical flow.
Once flows are predicted, they are directly applied to warp the first frame to synthesize future frames. 
Finally, a post-processing network is added to rectify imperfection caused during the warping operation. 
Inspired by\cite{meister2017unflow}, we further improve the performance of flow prediction and future frame generation using bidirectional flows and geometric consistency.
Experimental results demonstrate the effectiveness of the proposed method in video generation.

Our contributions are the following. 
\begin{enumerate}
    \item We introduce the novel task of conditioning video generation on a single semantic label map, allowing a good balance between flexibility and quality compared to existing video generation approaches. 
    \item 
    The difficult task is divided into two sub-problems, \ie, image generation followed by image-to-sequence generation, such that each stage can specialize on one problem.
    \item 
    We make full use of the semantic categorical prior in motion prediction when only one starting frame is available. It helps predict more accurate optical flow, thereby producing better future frames.
\end{enumerate}

\section{Related Work}
\label{sec:RelatedWorks}

{\bf Image generation} 
Many work exists regarding image generation which generally can be classified into two categories, unconditional generation and conditional generation. In unconditional generation, 
some work extends GANs \cite{goodfellow2014generative} or VAE \cite{kingma2013auto} to map from noise to real data distribution. Auto-regressive architectures model the image on a per-pixel basis \cite{uria2016neural, oord2016pixel}. 
In the second category, conditional models generate images given either class category, textual descriptions, scene graphs or images \cite{ma2017pose, balakrishnan2018synthesizing, zhang2017stackgan, johnson2018image, shrivastava2017learning}. Especially for image translation task, researchers study how to generate a meaningful image from a semantic representation such as semantic label maps (paired and unpaired) (\cite{isola2017image, zhu2017unpaired, wang2017high, bousmalis2017unsupervised, shrivastava2017learning}). However, in image generation tasks, photo-realism of the scene is modeled without considering their motion information.



{\bf Video Generation} 
Similar to Image generation, video generation can also be divided into two categories:
conditional and unconditional. For the former category, VideoGAN \cite{vondrick2016generating} explicitly disentangles a scene's foreground from background under the assumption that the background is stationary. The model is limited to only simple cases and cannot handle scenes with a moving background due to camera movement. TGAN \cite{saito2017temporal} first generates a sequence of latent variables and then synthesize a sequence of frames based on those latent variables. MoCoGAN \cite{tulyakov2017mocogan} also tries to map a sequence of random vectors to a sequence of frames. However, their framework decomposes video into content subspace and motion subspace, making video generation process more controllable. 
For conditional video generation, it is still at its early stage. One recent work is vid2vid~\cite{wang2018video} in which authors aim at transforming a sequence of semantic representation, e.g. semantic label map and sketch map, to a sequence of video frames. 
Our work falls into the category of conditional video generation, but unlike vid2vid, our method only requires a single semantic label map as input which enables more freedom over the generation process.

    
{\bf Video prediction} 
Some work model future motion in a deterministic manner. In \cite{ranzato2014video, srivastava2015unsupervised, villegas2017decomposing}, future prediction is carried out in a latent space, and the representation of future frames is projected back to image domain. These models are directly trained to optimize a reconstruction loss, such as Mean Squared Error (MSE), between the predicted frames and ground truth frames. However, they are prone to converging to blurry results as they compute an average of all possible future outcomes for the same starting frame. In \cite{luo2017unsupervised, jia2016dynamic, finn2016unsupervised}, future motion is predicted using either optical flow or filter, where estimation and then corresponding spatial transformation is applied to history frames to produce future frames. The result is sharp but lacks diversity.
%
A group of researchers \cite{xue2016visual, walker2017pose, denton2018stochastic, babaeizadeh2017stochastic}
introduced conditional variational autoencoders for video prediction to model uncertainty in future motion allowing the results to be both sharp and diverse.
%
Similar to our work, Walker et al. \cite{walker2016uncertain} and Li et al.~\cite{li2018flow} attempt to predict multiple future frames from a static image. In the training phase, they take the ground truth optical flow, either human annotated or computed, as supervision to predict such flow, and transform the given frame to future frames.
Contrary to Walker et al.~\cite{walker2016uncertain} and Li et al.~\cite{li2018flow}, we learn optical flow in an unsupervised manner, \ie, without taking any pre-computed flow as supervision. 
\section{Semantic Label Map to Video Generation}
\label{sec:Method}

%
Generating a video sequence $V=\{I_0, I_1, ...,I_T\}$ from a single semantic label map $S$ allows more flexibility compared to translating multiple label maps to a video, but is also more challenging. In this work we propose to divide such a difficult task into two relatively easy sub-problems and address each one separately, i.e., 
i) \textit{Image-to-Image} (I2I): an image generation model based on conditional GANs \cite{wang2017high} that maps a given semantic label map $S$ to the starting frame $\hat{I}_0$ of a sequence, and 
ii) \textit{Image-to-Video} (I2V): an image-sequence generation network that produces a sequence of frames $\hat{V}=\{ \hat{I}_0, \hat{I}_1, ...,\hat{I}_T\}$ based on the generated starting frame $\hat{I}_0$ and a latent variable $z$.
In each stage we have a model specializing on the corresponding task such that the overall performance is good.

\subsection{Image-to-Image (I2I)}
\label{subsec:img2img}
Image-to-image translation aims at learning the mapping of an image in the source domain to its corresponding image in the target domain. Among the existing methods \cite{isola2017image, zhu2017unpaired, wang2017high, bousmalis2017unsupervised}, we adopt the state-of-the-art image translation model pix2pixHD \cite{wang2017high} to generate an image from a semantic label map. It includes a coarse-to-fine architecture to progressively produce high quality images with fine details while keeping global consistency. Note that the translation stage is not restricted to this method and other image translation approaches can substitute pix2pixHD.

\subsection{Image-to-Video (I2V)} 
In this section, we present how to use cVAE for image sequence generation conditioned on an initial frame obtained from Sec.~\ref{subsec:img2img}.
It is composed of two sub-modules, i.e., flow prediction and video frame generation from flow.
Fig.~\ref{fig:architecture} shows the network structure and the components of the proposed Image-to-Video model. 
%

{\bf Conditional VAE - } 
Compared to future prediction from \textit{multiple} frames, where the future motion can be estimated based on past sequence, motion predicted from \textit{one} single frame can be more diverse. We employ the conditional VAE (cVAE) model~\cite{xue2016visual} as the backbone to capture multiple possible future motions conditioned on a static image. 
The proposed cVAE is composed of an encoder and a decoder.
The encoder $Q(z|V, I_0)$ learns to map a starting frame $I_0$ and the subsequent frames $V=\{I_1, ...,I_T\}$ into a latent variable $z$ that carries information about motion distribution conditioned on the first frame $I_0$. To achieve such mapping, the latent variable $z$ is composed of two parts, one projecting from the whole sequence including both $I_0$ and $V$, and the other from only the initial frame $I_0$.
The decoder $P(V|z,I_0)$ then reconstructs the sequence and outputs $\hat V$ based on a sampled $z$ and $I_0$. 
During training, the encoder $Q(z|V, I_0)$ learns to match the standard normal distribution, $N(0, I)$.
When running inference, the cVAE will generate a video sequence from a given starting frame $I_0$ and a latent variable $z$ sampled from $N(0, I)$ without the need of the motion encoder.

{\bf Flow Prediction - }  
%
We first use an image encoder to transform the starting frame into a latent vector $z_{I_0}$ as a part of the latent variable $z$.
The whole sequence is sent to another sequence encoder to compute $z_{m}$, which makes up the other part of $z$ for uncertainty modeling.
$z_{I_0}$ and $z_m$ are concatenated as one vector $z$ which is fed to a decoder to compute future optical flow.
For motion generation, we predict bidirectional flows, \ie both forward flow from the initial frame to future frames and backward flow from future frames to the initial frame. Computing cycle flow allows us to perform forward-backward consistency checks. 
For regions which appear in both frames (A and B), correspondence between two frames can be captured both from A to B and from B to A. We compute an occlusion mask to omit regions which are either occluded or missing in the generated frame so that the consistency check is only conducted on non-occluded regions. Putting all this together, the resulting output of the cVAE is the optical flow as well as the occlusion mask for both forward and backward directions, defined as: 
%
\begin{equation}
\begin{split}
W^f, W^b, O^f, O^b = \mathcal{F}(I_0),
\end{split}
\end{equation}
Where $\mathcal{F}$ is the flow prediction module that is composed of the motion encoder and the flow decoder as shown in Fig \ref{fig:architecture}. $W^f=\{{\bf w}_1^f,...,{\bf w}_T^f\}$, where ${\bf w}_t^f=(u^f, v^f)$ is the forward optical flow from $I_0$ to $I_t$ and  $W^b=\{{\bf w}_1^b,...,{\bf w}_T^b\}$, with ${\bf w}_t^b=(u^b, v^b)$ is the backward optical flow. $O^f =\{o_1^f,...,o_T^f\}$ and  $O^b =\{o_1^b,...,o_T^b\}$ are the multi-frame forward-backward occlusion maps. We define a pixel value in the occlusion map to be zero when there is no correspondence between frames. All optical flows and occlusion maps are jointly predicted by our image-to-flow module. Note that both bidirectional and occlusion maps are learned without any pre-computed flow as supervision. 

{\bf Video frame Generation - } 
With the predicted optical flow, we can directly produce future frames by warping the initial frame. However, the generated frames obtained solely by warping has inherent flaws, as some parts of the objects may not be visible in one frame but appears in another. To fill in the holes caused by either occlusion or objects entering or leaving the scene, we propose to add a post-processing network after frame warping. It takes a warped frame and its corresponding occlusion mask $O^b$ as the input, and generates the refined frame.
The final output of our model is defined as follows:
\begin{equation}
\begin{split}
\hat{I_t}({\bf x}) = \mathcal{P}(o_t^b({\bf x}) \cdot I_0({\bf x}+{\bf w}_t^b({\bf x}))),
\end{split}
\end{equation}
where $\mathcal{P}$ is the post-processing network 
and $\mathbf{x}$ denotes the coordinates of a position in the frame.

{\bf Loss Function - } 
Our loss function contains both per-pixel reconstruction and uncertainty modeling. For the per-pixel reconstruction, we compute losses in both the forward and backward direction, formulated as 
\begin{equation}
\begin{aligned}
\mathcal{L}_{r}(W^f, W^b, V) = \sum_{t}^T &\sum_{{\bf x}} o_t^f({\bf x}) |I_0({\bf x})-I_t({\bf x}+{\bf w}^f_t({\bf x}))|_1 \\
& + o_t^b({\bf x}) |I_t({\bf x})-I_0({\bf x}+{\bf w}^b_t({\bf x}))|_1 ,
\end{aligned}
\end{equation}
where $T$ is the length of the generated sequence. 
We only compute reconstruction in non-occluded regions to avoid learning incorrect deformations. Neighboring pixels usually belong to the same object, thus they tend to have similar displacement. Therefore, similar to previous work \cite{zhang2014rigid, trobin2008unbiased} we also add a smoothness constraint to encourage flow in a local neighborhood to be similar. 
\begin{equation}
\mathcal{L}_{fs}(W^f, W^b) = |\nabla W^f|_1 + |\nabla W^b|_1
\end{equation}
We compute forward-backward consistency loss for non-occluded regions:
\begin{equation}
\begin{aligned}
\mathcal{L}_{fc}(W^f, W^b) = \sum_{t}^T &\sum_{{\bf x}} o_t^f({\bf x}) |{\bf w}^f_t({\bf x})-{\bf w}^b_t({\bf x}+{\bf w}^f_t({\bf x}))|_1 \\
 & + o_t^b({\bf x}) |{\bf w}^b_t({\bf x})-{\bf w}^f_t({\bf x}+{\bf w}^b_t({\bf x}))|_1,
\end{aligned}
\end{equation}

To train the in-painting network, we applied an $L1$ loss together with a perceptual loss \cite{johnson2016perceptual} that has been shown to be useful for image generation. Therefore, our data loss can be formulated as a weighed sum of the above terms.
\begin{equation}
\begin{aligned}
\mathcal{L}_{data}(\hat V, V) &= \lambda _r \mathcal{L}_{r} + \lambda _{fs} \mathcal{L}_{fs} + \lambda _{fc}\mathcal{L}_{fc}  \\
&+ \mathcal{L}_{l1}(\hat{V},V)+ \mathcal{L}_{l1}(\phi(\hat{V}), \phi(V)) \\
&+ \lambda_p |1-O^b |_1 + \lambda_p |1-O^f |_1,
\end{aligned}
\end{equation}
where $\phi$ is VGG-19 \cite{Simonyan14c} from where we extract and collect features from the first 16 layers. We add a penalty on the occlusion maps for $\lambda_p=0.1$ to avoid the trivial solution where all pixels become occluded (we define the value in a position of $O^b$ to be $0$ when the pixels is becoming occluded in the next frame). The weights are set to be: $\lambda_r=\lambda_{fs}=\lambda_{fc}=\lambda_{l1}=1$ and $\beta=0.1$.
To model the motion uncertainty we incorporate the KL-divergence loss such that $Q(z|X)$ matches $N(0, I)$. The training loss for the cVAE is a data loss combined with a KL-divergence loss.
\begin{equation}
\begin{aligned}
\mathcal{L}_{cVAE}( \hat{V}, V) =& \mathcal{L}_{data} + \beta \mathcal{D}_{kl}(p_\phi(z|V) || p(z)).
\end{aligned}
\end{equation}

\subsection{Flow prediction with semantic label maps} 
Different from video prediction conditioned on multiple frames, generating a video from a static frame has no access to historical motion information. To infer future motion of a object in a static frame, the model needs to understand the semantic category of that object and its interaction with other objects and background. For example, the car will stop when the traffic light is red and move on when is green. To promote future motion estimation for the whole frame, we incorporate semantic label map which describes semantic information of the whole scene into the flow prediction module discussed in previous sub-section.

\begin{figure}[t]
\includegraphics[width=\linewidth]{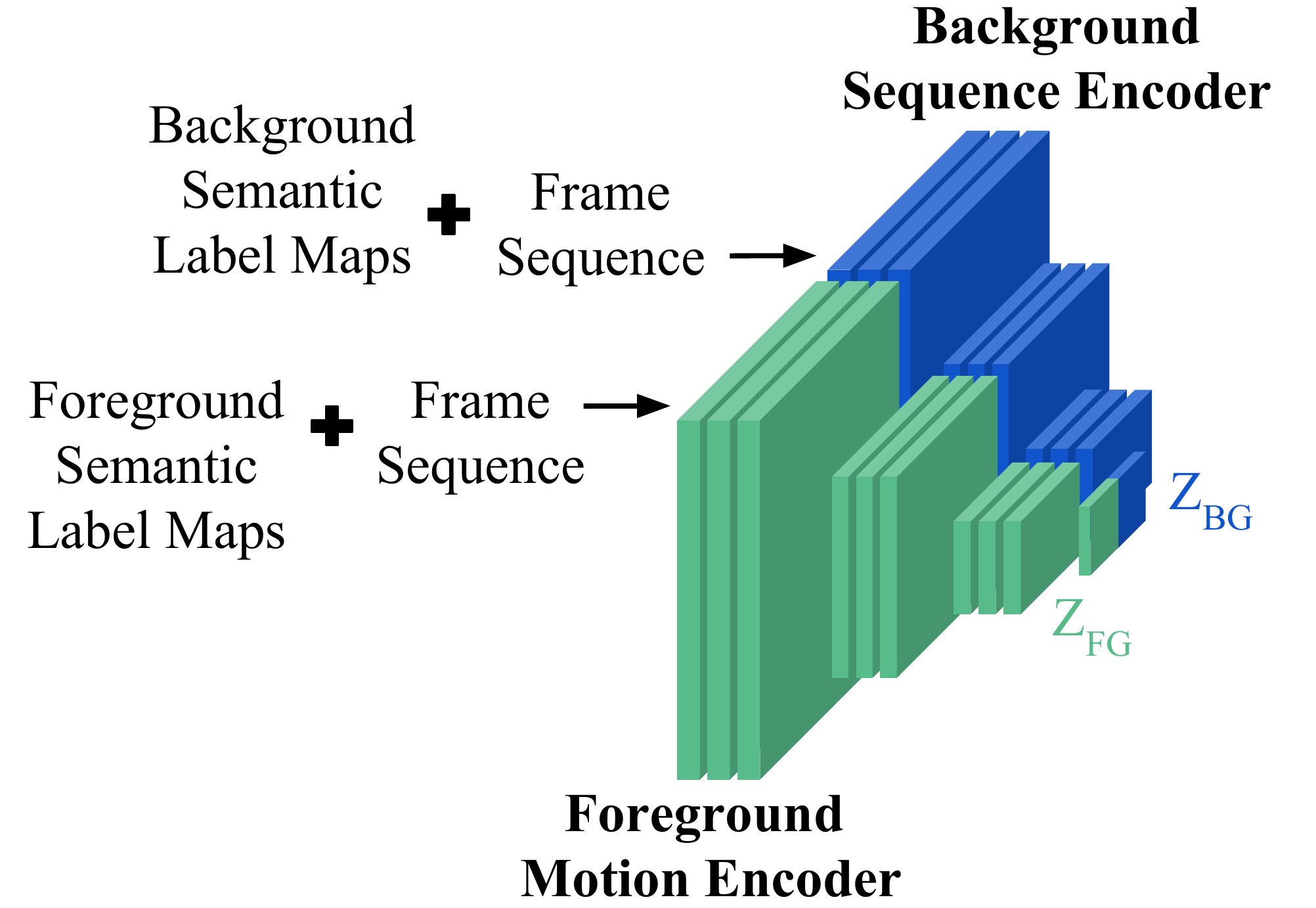}
\caption{Semantic sequence encoder. Each sequence encoder only focuses on learning either foreground or background motion.}
\label{fig:semantic_motion_encoder}
\end{figure}

We explore two ways of integrating the semantic label map for flow prediction. In the first method, we expand a semantic label map into several heatmaps which is filled with ones on positions correspond to a semantic category and zeros elsewhere. These heatmaps are concatenated with the generated starting frame and fed to the cVAE model for future frame synthesis. In the other method, we further divide the heatmaps into two sets, \ie, foreground heatmaps and background heatmaps, as shown in Fig.~\ref{fig:semantic_motion_encoder}. Each set of heatmaps is fed to a separate sequence encoder to get a latent vector $z_{FG}$ and $z_{BG}$. They are then concatenated with $z_{I_0}$ becoming the input to the flow decoder. In Section~\ref{sec:Experiments}, experimental results demonstrate that integrating semantic label map helps computing more accurate flow and accordingly improve the video generation performance.


\section{Experiments}
\label{sec:Experiments}
In this section we present the dataset and describe the details about the implementation. We evaluate our method against several baseline methods with both qualitative and quantitative metrics.  We also perform ablation studies to confirm the effectiveness of using semantic label maps for video generation.

\subsection{Datasets and Evaluation Metrics}

{\bf Datasets} We have conducted experiments on the Cityscapes dataset while we have provided qualitative results on the many other datasets. {\bf Cityscapes} \cite{Cordts2016Cityscapes} consists of urban scene videos recorded from a car driving on the street. It contains 2,975 training, 500 validation and 1,525 test video sequences, each containing 30 frames. The ground truth semantic segmentation mask is only available for the 20th frame of every video. We use DeepLabV3\cite{deeplabv3plus2018} to compute  semantic segmentation maps for all frames, which are used for training and testing. We train the model using all videos from the training set, and test it on the validation set. {\bf UCF101} \cite{soomro2012ucf101} The dataset contains $13,220$ videos of $101$ action classes. {\bf KTH Action dataset} \cite{laptev2004recognizing} consists of 600 videos of people performing one of the six actions(walking, jogging, running, boxing, handwaving, hand-clapping). {\bf KITTI} \cite{Geiger2013IJRR} similar to  Cityscpes  was recorded from a car traversing streets. 

{\bf Evaluation Metrics} 
We provide both quantitative and qualitative evaluation results in this section.
For qualitative evaluation, we conducted a human subjective study to  evaluate our method as well as the baseline methods. We randomly generated 100 video sequences for each method, pairing each generated video with the result of another randomly chosen method. The participants are asked to choose from each pair the most realistic looking video. We calculate the human preference score after each pair of videos was evaluated by 10 participants.

The Fr\'echet Inception Distance (FID) \cite{heusel2017gans} measures the similarity between two sets of images.  It was shown to correlate well with human judgment of visual quality and is most often used to evaluate the quality of samples from GANs. FID is calculated by computing the Fr\'echet distance between two feature representations of the Inception network.  Similar to \cite{wang2018video}, we use the video inception network \cite{Carreira_2017_CVPR} to extract spatio-temporal feature representations.

\subsection{Implementation details}
Our method takes a single semantic label map $S$ and predict $T$ = 8 frames in a single step. We resize all frames to $128\times128$ and extract the semantic segmentation maps with DeepLabV3 \cite{deeplabv3plus2018} for training. We do not use any flow map as ground truth for training. In the cVAE, the motion encoder is 
built upon stacks of 2D convolutional layers intercepted with max pooling layers. The latent vector $z$ has dimension 1024, 896 for foreground motion and 128 for background motion. For the flow encoder, we use three blocks each consisting of 3D convolutional layers intercepted with bilinear upsampling layer that progressively recovers the input resolution in both spatial and temporal dimensions. For the postprocessing network, we adopt the U-Net architecture from \cite{ronneberger2015u}. 

\subsection{Ablation Studies}
\begin{figure}[t]
\vspace{-2mm}
\includegraphics[width=\linewidth]{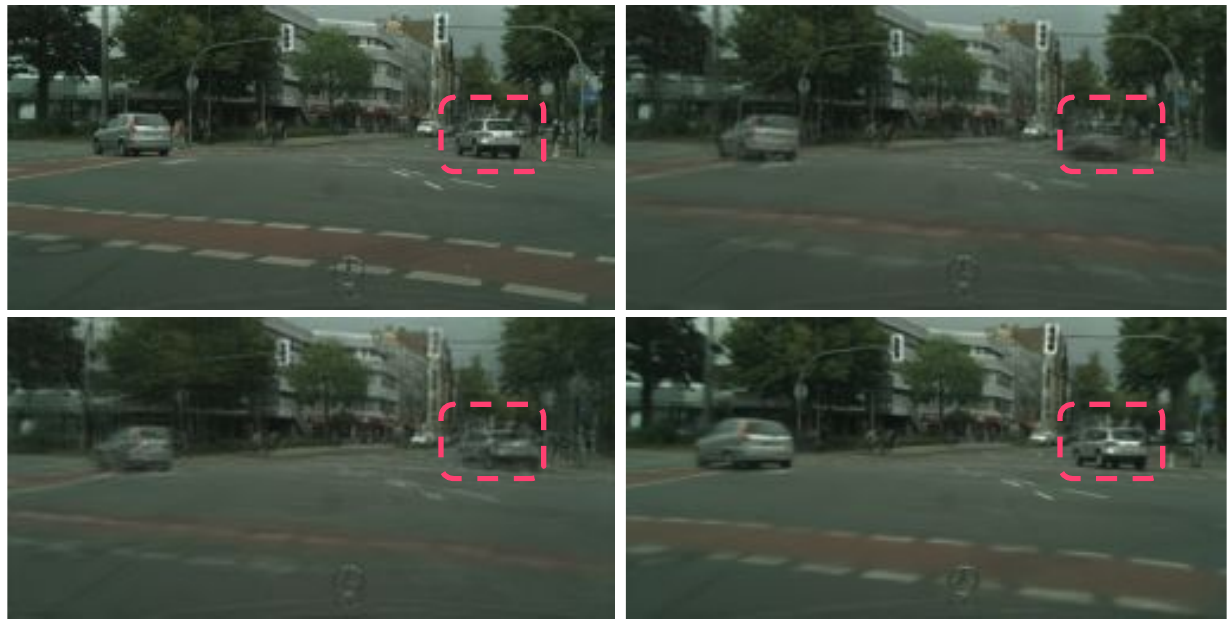}
\caption{Ablation studies of our method. Top left: GT. Top right: w/o segmentation label map and flow. Bottom left:w/o flow. Bottom right: our full model. Our method preserve better the visual quality.}
\label{fig:ablation_rebuttal}
\end{figure}

We conduct extensive experiments on the Cityscapes dataset to analyze the contribution of the semantic label map and optical flow for motion prediction. We have shown that optical flow is reliable motion representation to convey motion between frames and preserver better visual quality. Fig. ~\ref{fig:ablation_rebuttal} shows that the model without optical flow produces blurry frames. In contrast, our flow based solution preserves better details even on fast moving objects and produces fewer artifacts.

\begin{figure}[t]
\vspace{-3mm}
\includegraphics[width=\linewidth]{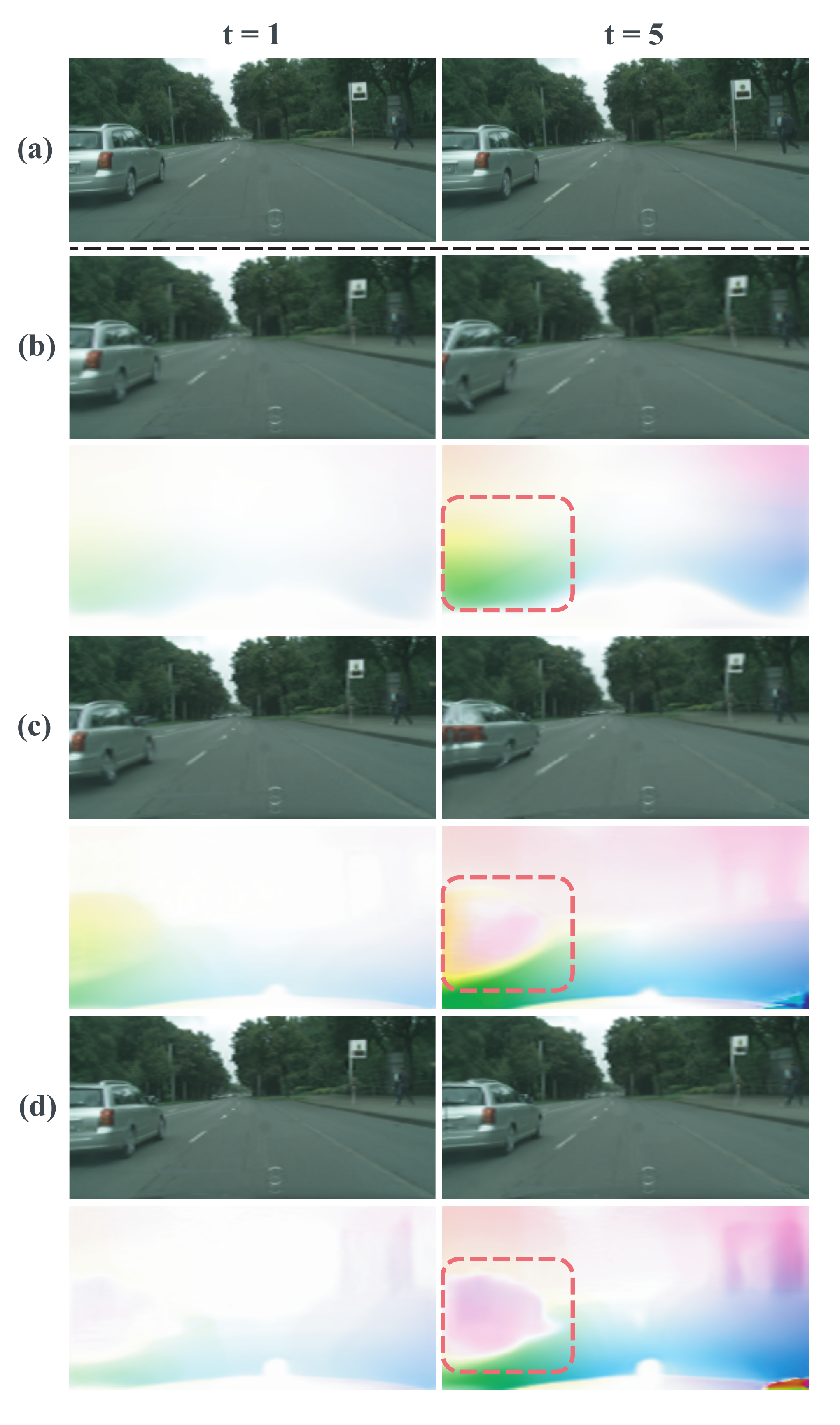}
\caption{We compare three different variants of using semantic label map for flow and frame prediction. (a) ground truth, (b) w/o semantic label maps, (c) with semantic label maps, (d) with separate semantic label maps for background and foreground objects.}
\vspace{-3mm}
\label{fig:ablation}
\end{figure}

We also compare frame sequences generated by the model without semantic label map and two ways of integrating that.
As shown in Fig.~\ref{fig:ablation}, the model integrating semantic label map is able to capture both foreground object motion and background motion, whereas the one without that fails to estimate the independent foreground object motion. By further separating semantic label maps into background and foreground, it can capture more details in structure marked by the red rectangles. As expected, semantic information plays an important role in generating object motion when predicting from a single frame. We show further improvements by separating semantic classes into two groups based on background and foreground. 
\begin{figure}[t]
\includegraphics[width=\linewidth]{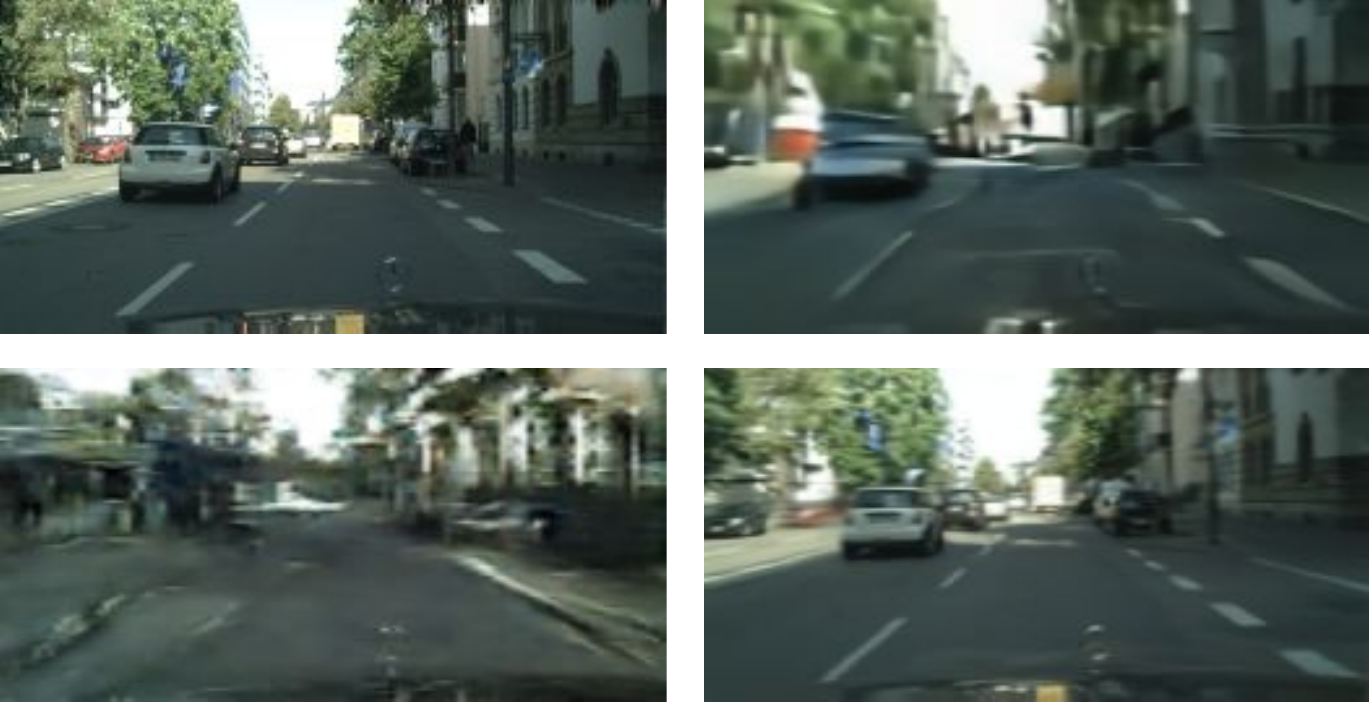}
\caption{Comparison between different approaches of video prediction from a static image. Top left: ground truth. Top right: FG. Bottom left: MoCoGAN. Bottom right: img2vid (ours). Our method preserve the the visual quality while other method rapidly degrades.}
\label{fig:prediction_comparisons}
\end{figure}

\begin{figure*}[!ht]
\includegraphics[width=\textwidth]{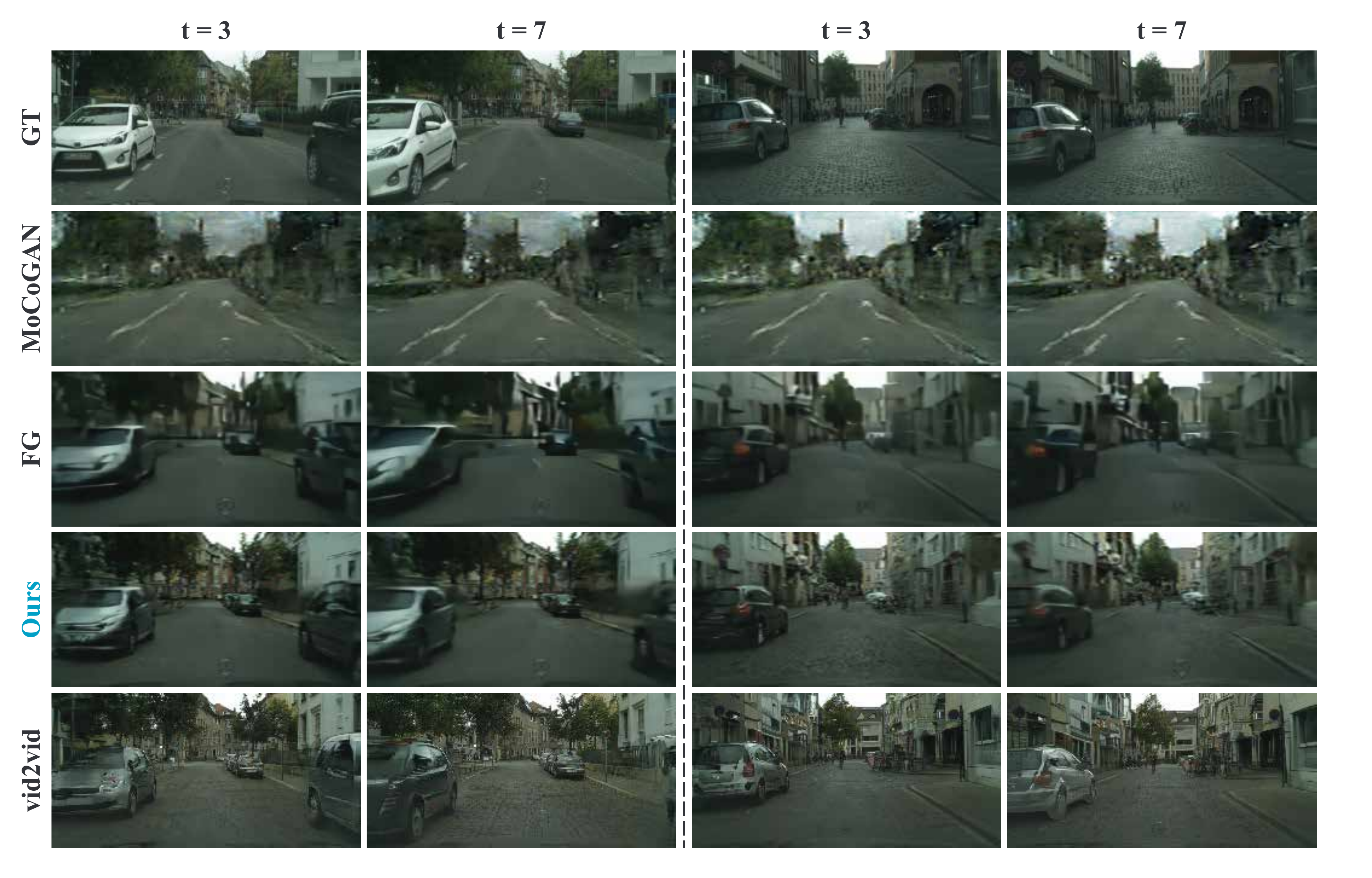}
\caption{Comparisons with other competing baselines. Notice that vid2vid uses a sequence of semantic label maps while other methods only take {\bf one} as input. Please zoom in for best view.}
\label{fig:generation_comparison}
\end{figure*}

\begin{table}[]
\centering
\begin{tabular}{cccccc}
\hline
     & MoCoGAN & FG   & vid2vid & Ours          \\ \hline
FID  & 8.77    & 3.69 & 4.86    & \textbf{3.52}
\\ \hline
\end{tabular}
\caption{Comparison of video generation methods where the input is a single semantic label map.
}
\label{table:fid}
\end{table}

\begin{table}[]
\centering
\begin{tabular}{cccccc}
\hline
    & MoCoGAN    & FG      & Ours          \\ \hline
FID & 7.06      & 2.86    & \textbf{1.80}
\\ \hline
\end{tabular}
\caption{Comparison of video prediction methods that take a single starting frame as input.
}
\label{table:fid_prediction}
\end{table}

\subsection{Baselines}
We compare our network with five state-of-the-art baseline methods trained on the Cityscapes dataset.

{\bf MoCoGAN} \cite{tulyakov2017mocogan} is an unconditional video generation model. Here, we also compared the conditional setting of MoCoGAN, given the initial frame $x_0$ as input.

{\bf FlowGrounded (FG)} \cite{li2018flow} is a video prediction model  from a static image. We compare our image-to-video stage with this method on both video generation and video prediction tasks.


{\bf Vid2Vid} \cite{wang2018video}, the goal of vid2vid is to map a sequence of semantic represenation to a sequence of video frames, where future motion is approximately given in the semantic segmentation sequence.
We evaluate vid2vid to see whether our method is comparable to this "upper bound".


\begin{figure}[t]
\includegraphics[width=\linewidth]{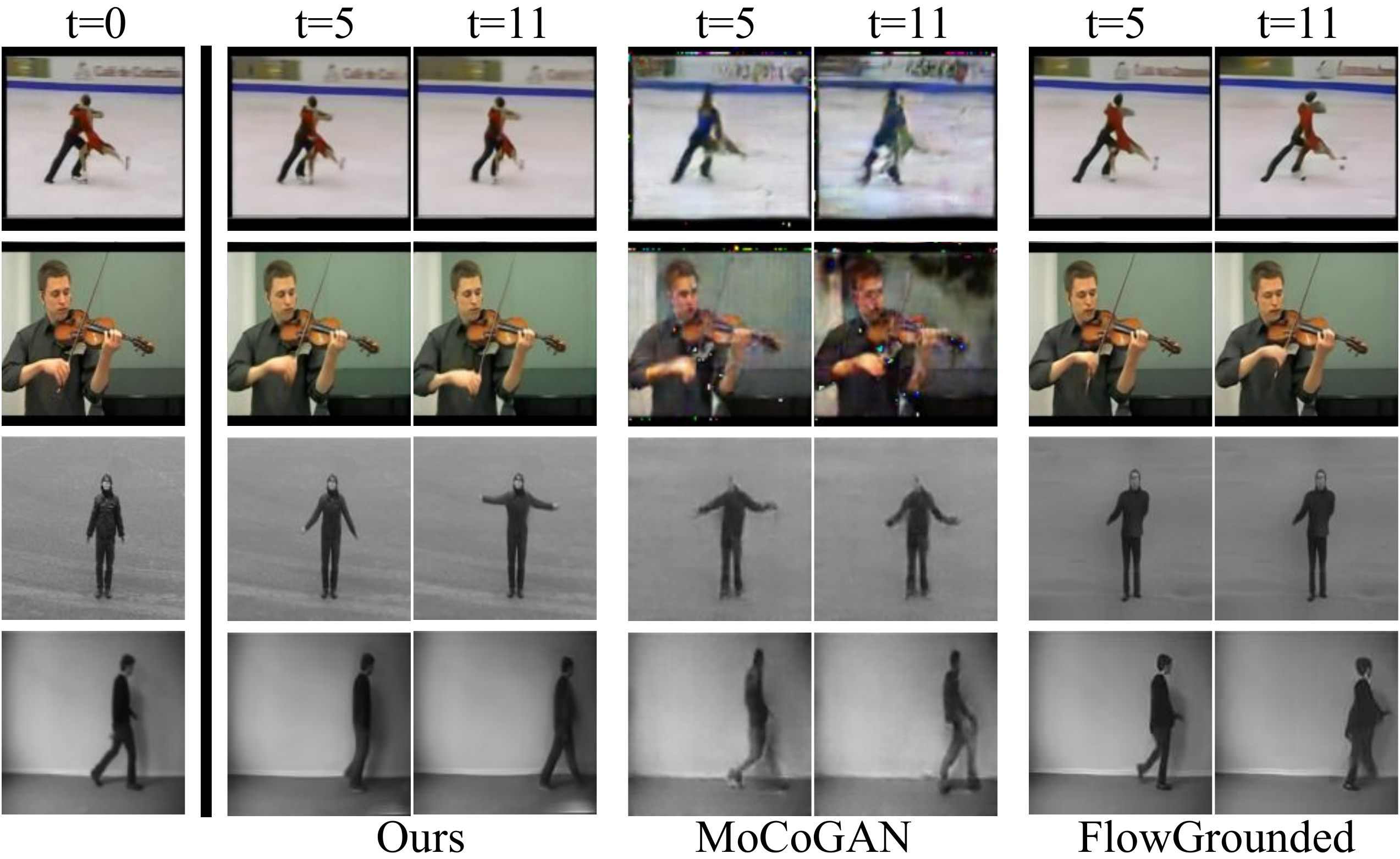}
\caption{Comparisons with other competing baselines on UCF-101 dataset and KTH human dataset. Please zoom in to see the details.}
\vspace{-5mm}
\label{fig:generation_comparison_rebuttal}
\end{figure}

\begin{figure}[t]
\vspace{-2mm}
\includegraphics[width=\linewidth]{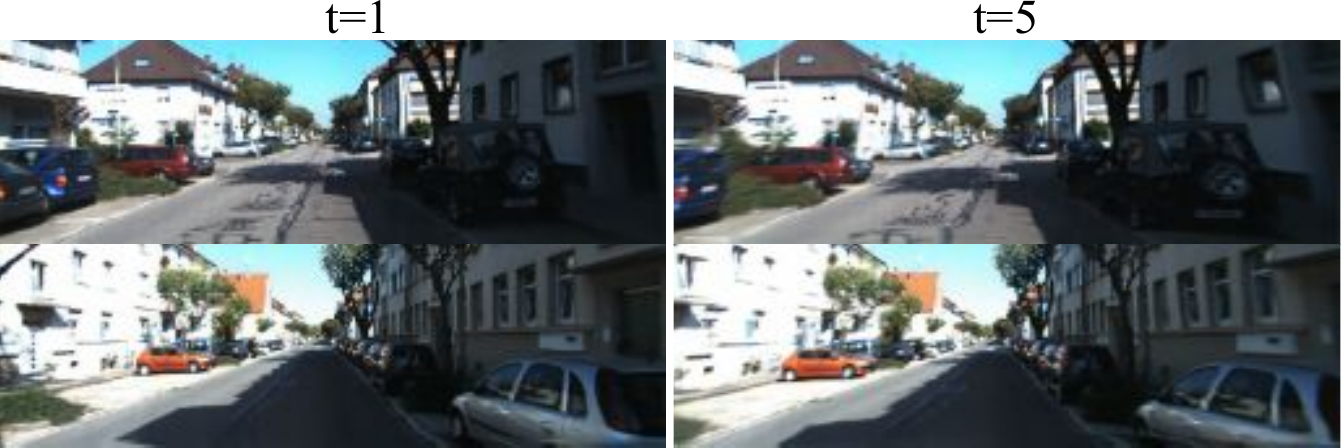}
\caption{Samples of KITTI generated from model trained on the cityscapes dataset.}
\vspace{-3mm}
\label{fig:kitti}
\end{figure}

\subsection{Results}

{\bf Quantitative Results} In Table~\ref{table:fid} we report the results on the Cityscapes dataset. In terms of performance, the lower the FID, the better the model. In Table~\ref{table:fid}, we show that our method has the lowest FID compared to all competing methods. Notice that the results here are slightly different from what is reported by Wang et al. \cite{wang2017high} because we only evaluate 8-frame sequences with a resolution of $1024 \times 512$ due to GPU memory limitations. We generated a total of 500 short sequences on the validation set. We also provide results for video prediction when only the starting frame is given. As shown in Table ~\ref{table:fid_prediction}, our method outperforms all other state-of-the-art approaches in video prediction from a static image.

{\bf Qualitative Results}  Fig.~\ref{fig:generation_comparison} compares our generation results with other approaches.
MoCoGAN has limited capability in modeling video sequences (both motion and appearance). FG fails to synthesize the details of the scene,\eg windows of the background building are completely missing, increasing blurriness. Our method maintains the semantic structure of the scene for the duration of the sequence and contains finer details than the previous two methods. The proposed method makes reasonable estimates of the objects' future motion and produces temporally coherent video sequence. Compared to the ground truth sequence, our model can generate semantically correct samples but with different properties, \eg, a white car in the ground truth sequence appears as a silver car in our result. For vid2vid, where the input is a sequence of semantic label maps, shows realistic images with great details, but limited on preserving the temporal consistency across frames, \eg the silver car in $t=3$ has turned into black in $t=7$, while our methods keeps the same color. 
To further show the effectiveness of our method on predicting general motions, we provide visual results on UCF-101 dataset and KTH action dataset that mainly consist on people performing actions. As shown in Fig.~\ref{fig:generation_comparison_rebuttal}, our method preserves well the body structure and synthesizing complex non-linear motions such as people skiing, playing violin and walking. We trained the model on Cityscapes and tested on samples from KITTI to show the method`s generalization ability, shown in Fig.~\ref{fig:kitti}.
\begin{table}[]
\centering
\begin{tabular}{cc}
\hline
\multicolumn{2}{c}{Human Preference Score}                                 \\ \hline
seg2vid(ours) / MoCoGAN                       & {\bf 1.0} / 0.0                 \\
seg2vid(ours) / FG                            & {\bf 0.78} / 0.22           \\
seg2vid(ours) /vid2vid                        & 0.37 / {\bf 0.63}             \\ \hline
\end{tabular}
\caption{User study on video generation methods.}
\label{table:human_preference}
\end{table}
\begin{table}[]
\centering
\begin{tabular}{cc}
\hline
\multicolumn{2}{c}{Human Preference Score}                                     \\ \hline
seg2vid(ours) / MoCoGAN                       & {\bf 1.0} / 0.0                     \\
seg2vid(ours) / FG                            & {\bf 0.82} / 0.18 \\ \hline
\end{tabular}
\caption{User study on video prediction methods.}
\vspace{-5mm}
\label{table:human_preference_prediction}
\end{table}

The user study illustrated in Table. ~\ref{table:human_preference} also shown that our method is the most favored except vid2vid.  Additionally to the results of synthesized data, we also reported results for video prediction task. 
As shown in Fig. ~\ref{fig:prediction_comparisons} our method can predict well background motion and simultaneously captured the movement of the car on the left side. The details and structure of the scene is well preserved with our approach while other methods suffer severe deformation. Table \ref{table:human_preference_prediction} shows that participants find our method to be more realistic.

\section{Conclusion}
\label{sec:Conclusion}

 
In this work, we introduced the new video generation task conditioned only on a single semantic label map, and proposed a novel method for this task. Instead of learning the generation end-to-end, which is very challenging, we employed a divide and conquer strategy to model appearance and motion in a progressive manner to obtain quality results. We demonstrated that introducing semantic information brings large improvement when predicting motion from static content. The impressive performance compared to other baselines indicate the effectiveness of the proposed method for video generation. 


{\small
\bibliographystyle{ieee}
\bibliography{egbib}

\begin{thebibliography}{10}\itemsep=-1pt

\bibitem{babaeizadeh2017stochastic}
M.~Babaeizadeh, C.~Finn, D.~Erhan, R.~H. Campbell, and S.~Levine.
\newblock Stochastic variational video prediction.
\newblock {\em arXiv preprint arXiv:1710.11252}, 2017.

\bibitem{balakrishnan2018synthesizing}
G.~Balakrishnan, A.~Zhao, A.~V. Dalca, F.~Durand, and J.~Guttag.
\newblock Synthesizing images of humans in unseen poses.
\newblock {\em arXiv preprint arXiv:1804.07739}, 2018.

\bibitem{bousmalis2017unsupervised}
K.~Bousmalis, N.~Silberman, D.~Dohan, D.~Erhan, and D.~Krishnan.
\newblock Unsupervised pixel-level domain adaptation with generative
  adversarial networks.
\newblock In {\em The IEEE Conference on Computer Vision and Pattern
  Recognition (CVPR)}, volume~1, page~7, 2017.

\bibitem{Carreira_2017_CVPR}
J.~Carreira and A.~Zisserman.
\newblock Quo vadis, action recognition? a new model and the kinetics dataset.
\newblock In {\em The IEEE Conference on Computer Vision and Pattern
  Recognition (CVPR)}, July 2017.

\bibitem{deeplabv3plus2018}
L.-C. Chen, Y.~Zhu, G.~Papandreou, F.~Schroff, and H.~Adam.
\newblock Encoder-decoder with atrous separable convolution for semantic image
  segmentation.
\newblock In {\em ECCV}, 2018.

\bibitem{Cordts2016Cityscapes}
M.~Cordts, M.~Omran, S.~Ramos, T.~Rehfeld, M.~Enzweiler, R.~Benenson,
  U.~Franke, S.~Roth, and B.~Schiele.
\newblock The cityscapes dataset for semantic urban scene understanding.
\newblock In {\em Proc. of the IEEE Conference on Computer Vision and Pattern
  Recognition (CVPR)}, 2016.

\bibitem{denton2018stochastic}
E.~Denton and R.~Fergus.
\newblock Stochastic video generation with a learned prior.
\newblock {\em arXiv preprint arXiv:1802.07687}, 2018.

\bibitem{finn2016unsupervised}
C.~Finn, I.~Goodfellow, and S.~Levine.
\newblock Unsupervised learning for physical interaction through video
  prediction.
\newblock In {\em Advances in neural information processing systems}, pages
  64--72, 2016.

\bibitem{Geiger2013IJRR}
A.~Geiger, P.~Lenz, C.~Stiller, and R.~Urtasun.
\newblock Vision meets robotics: The kitti dataset.
\newblock {\em International Journal of Robotics Research (IJRR)}, 2013.

\bibitem{goodfellow2014generative}
I.~Goodfellow, J.~Pouget-Abadie, M.~Mirza, B.~Xu, D.~Warde-Farley, S.~Ozair,
  A.~Courville, and Y.~Bengio.
\newblock Generative adversarial nets.
\newblock In {\em Advances in neural information processing systems}, pages
  2672--2680, 2014.

\bibitem{heusel2017gans}
M.~Heusel, H.~Ramsauer, T.~Unterthiner, B.~Nessler, and S.~Hochreiter.
\newblock Gans trained by a two time-scale update rule converge to a local nash
  equilibrium.
\newblock In {\em Advances in Neural Information Processing Systems}, pages
  6626--6637, 2017.

\bibitem{isola2017image}
P.~Isola, J.-Y. Zhu, T.~Zhou, and A.~A. Efros.
\newblock Image-to-image translation with conditional adversarial networks.
\newblock {\em arXiv preprint}, 2017.

\bibitem{jia2016dynamic}
X.~Jia, B.~De~Brabandere, T.~Tuytelaars, and L.~V. Gool.
\newblock Dynamic filter networks.
\newblock In {\em Advances in Neural Information Processing Systems}, pages
  667--675, 2016.

\bibitem{johnson2016perceptual}
J.~Johnson, A.~Alahi, and L.~Fei-Fei.
\newblock Perceptual losses for real-time style transfer and super-resolution.
\newblock In {\em European Conference on Computer Vision}, pages 694--711.
  Springer, 2016.

\bibitem{johnson2018image}
J.~Johnson, A.~Gupta, and L.~Fei-Fei.
\newblock Image generation from scene graphs.
\newblock In {\em CVPR}, 2018.

\bibitem{kingma2013auto}
D.~P. Kingma and M.~Welling.
\newblock Auto-encoding variational bayes.
\newblock {\em arXiv preprint arXiv:1312.6114}, 2013.

\bibitem{laptev2004recognizing}
I.~Laptev, B.~Caputo, et~al.
\newblock Recognizing human actions: a local svm approach.
\newblock In {\em null}, pages 32--36. IEEE, 2004.

\bibitem{li2018flow}
Y.~Li, C.~Fang, J.~Yang, Z.~Wang, X.~Lu, and M.-H. Yang.
\newblock Flow-grounded spatial-temporal video prediction from still images.
\newblock {\em arXiv preprint arXiv:1807.09755}, 2018.

\bibitem{luo2017unsupervised}
Z.~Luo, B.~Peng, D.-A. Huang, A.~Alahi, and L.~Fei-Fei.
\newblock Unsupervised learning of long-term motion dynamics for videos.
\newblock {\em arXiv preprint arXiv:1701.01821}, 2, 2017.

\bibitem{ma2017pose}
L.~Ma, X.~Jia, Q.~Sun, B.~Schiele, T.~Tuytelaars, and L.~Van~Gool.
\newblock Pose guided person image generation.
\newblock In {\em Advances in Neural Information Processing Systems}, pages
  406--416, 2017.

\bibitem{meister2017unflow}
S.~Meister, J.~Hur, and S.~Roth.
\newblock Unflow: Unsupervised learning of optical flow with a bidirectional
  census loss.
\newblock {\em arXiv preprint arXiv:1711.07837}, 2017.

\bibitem{oord2016pixel}
A.~v.~d. Oord, N.~Kalchbrenner, and K.~Kavukcuoglu.
\newblock Pixel recurrent neural networks.
\newblock {\em arXiv preprint arXiv:1601.06759}, 2016.

\bibitem{ranzato2014video}
M.~Ranzato, A.~Szlam, J.~Bruna, M.~Mathieu, R.~Collobert, and S.~Chopra.
\newblock Video (language) modeling: a baseline for generative models of
  natural videos.
\newblock {\em arXiv preprint arXiv:1412.6604}, 2014.

\bibitem{ronneberger2015u}
O.~Ronneberger, P.~Fischer, and T.~Brox.
\newblock U-net: Convolutional networks for biomedical image segmentation.
\newblock In {\em International Conference on Medical image computing and
  computer-assisted intervention}, pages 234--241. Springer, 2015.

\bibitem{saito2017temporal}
M.~Saito, E.~Matsumoto, and S.~Saito.
\newblock Temporal generative adversarial nets with singular value clipping.
\newblock In {\em IEEE International Conference on Computer Vision (ICCV)},
  volume~2, page~5, 2017.

\bibitem{shrivastava2017learning}
A.~Shrivastava, T.~Pfister, O.~Tuzel, J.~Susskind, W.~Wang, and R.~Webb.
\newblock Learning from simulated and unsupervised images through adversarial
  training.
\newblock In {\em CVPR}, volume~2, page~5, 2017.

\bibitem{Simonyan14c}
K.~Simonyan and A.~Zisserman.
\newblock Very deep convolutional networks for large-scale image recognition.
\newblock {\em CoRR}, abs/1409.1556, 2014.

\bibitem{soomro2012ucf101}
K.~Soomro, A.~R. Zamir, and M.~Shah.
\newblock Ucf101: A dataset of 101 human actions classes from videos in the
  wild.
\newblock {\em arXiv preprint arXiv:1212.0402}, 2012.

\bibitem{srivastava2015unsupervised}
N.~Srivastava, E.~Mansimov, and R.~Salakhudinov.
\newblock Unsupervised learning of video representations using lstms.
\newblock In {\em International conference on machine learning}, pages
  843--852, 2015.

\bibitem{trobin2008unbiased}
W.~Trobin, T.~Pock, D.~Cremers, and H.~Bischof.
\newblock An unbiased second-order prior for high-accuracy motion estimation.
\newblock In {\em Joint Pattern Recognition Symposium}, pages 396--405.
  Springer, 2008.

\bibitem{tulyakov2017mocogan}
S.~Tulyakov, M.-Y. Liu, X.~Yang, and J.~Kautz.
\newblock Mocogan: Decomposing motion and content for video generation.
\newblock {\em arXiv preprint arXiv:1707.04993}, 2017.

\bibitem{uria2016neural}
B.~Uria, M.-A. C{\^o}t{\'e}, K.~Gregor, I.~Murray, and H.~Larochelle.
\newblock Neural autoregressive distribution estimation.
\newblock {\em The Journal of Machine Learning Research}, 17(1):7184--7220,
  2016.

\bibitem{villegas2017decomposing}
R.~Villegas, J.~Yang, S.~Hong, X.~Lin, and H.~Lee.
\newblock Decomposing motion and content for natural video sequence prediction.
\newblock {\em arXiv preprint arXiv:1706.08033}, 2017.

\bibitem{vondrick2016generating}
C.~Vondrick, H.~Pirsiavash, and A.~Torralba.
\newblock Generating videos with scene dynamics.
\newblock In {\em Advances In Neural Information Processing Systems}, pages
  613--621, 2016.

\bibitem{walker2016uncertain}
J.~Walker, C.~Doersch, A.~Gupta, and M.~Hebert.
\newblock An uncertain future: Forecasting from static images using variational
  autoencoders.
\newblock In {\em European Conference on Computer Vision}, pages 835--851.
  Springer, 2016.

\bibitem{walker2017pose}
J.~Walker, K.~Marino, A.~Gupta, and M.~Hebert.
\newblock The pose knows: Video forecasting by generating pose futures.
\newblock In {\em Computer Vision (ICCV), 2017 IEEE International Conference
  on}, pages 3352--3361. IEEE, 2017.

\bibitem{wang2018video}
T.-C. Wang, M.-Y. Liu, J.-Y. Zhu, G.~Liu, A.~Tao, J.~Kautz, and B.~Catanzaro.
\newblock Video-to-video synthesis.
\newblock {\em arXiv preprint arXiv:1808.06601}, 2018.

\bibitem{wang2017high}
T.-C. Wang, M.-Y. Liu, J.-Y. Zhu, A.~Tao, J.~Kautz, and B.~Catanzaro.
\newblock High-resolution image synthesis and semantic manipulation with
  conditional gans.

\bibitem{xue2016visual}
T.~Xue, J.~Wu, K.~Bouman, and B.~Freeman.
\newblock Visual dynamics: Probabilistic future frame synthesis via cross
  convolutional networks.
\newblock In {\em Advances in Neural Information Processing Systems}, pages
  91--99, 2016.

\bibitem{zhang2014rigid}
C.~Zhang, Z.~Li, R.~Cai, H.~Chao, and Y.~Rui.
\newblock As-rigid-as-possible stereo under second order smoothness priors.
\newblock In {\em European Conference on Computer Vision}, pages 112--126.
  Springer, 2014.

\bibitem{zhang2017stackgan}
H.~Zhang, T.~Xu, H.~Li, S.~Zhang, X.~Huang, X.~Wang, and D.~Metaxas.
\newblock Stackgan: Text to photo-realistic image synthesis with stacked
  generative adversarial networks.
\newblock {\em arXiv preprint}, 2017.

\bibitem{zhu2017unpaired}
J.-Y. Zhu, T.~Park, P.~Isola, and A.~A. Efros.
\newblock Unpaired image-to-image translation using cycle-consistent
  adversarial networks.
\newblock {\em arXiv preprint}, 2017.

\end{thebibliography}
}
\end{document}